\documentclass[10pt,twocolumn,letterpaper]{article}

\usepackage[pagenumbers]{wacv} 
\makeatother

\def\mE{\mathcal{E}}
\def\mF{\mathcal{F}}

\def\mL{\mathcal{L}}

\def\1n{\mathbf{1}_n}
\def\0{\mathbf{0}}
\def\1{\mathbf{1}}

\definecolor{wacvblue}{rgb}{0.21,0.49,0.74}
\usepackage[pagebackref,breaklinks,colorlinks,allcolors=wacvblue]{hyperref}
\usepackage[table]{xcolor}
\usepackage[dvipsnames]{xcolor}
\def\wacvPaperID{1451} 
\def\confName{WACV}
\def\confYear{2026}

\title{CORA: Consistency-Guided Semi-Supervised Framework for\\ Reasoning Segmentation}

\author{
Prantik Howlader$^{\dagger}$ \quad
Hoang Nguyen\textnormal{-}Canh$^{\dagger}$ \quad
Srijan Das$^{\ddagger}$ \quad
Jingyi Xu$^{\dagger}$ \quad
Hieu Le$^{\ddagger}$ \quad
Dimitris Samaras$^{\dagger}$ \\[0.4em]
$^{\dagger}$Stony Brook University \quad
$^{\ddagger}$UNC-Charlotte \\[0.25em]
{\tt\small
\{phowlader,hcnguyen,jingyixu,samaras\}@cs.stonybrook.edu \quad
\{sdas24,hle40\}@charlotte.edu 
}
}

\begin{document}
\maketitle
\newcommand{\rev}[1]{{\color{blue} #1}}
\begin{abstract}
Reasoning segmentation seeks pixel-accurate masks for targets referenced by complex, often implicit instructions, requiring context-dependent reasoning over the scene. Recent multimodal language models have advanced instruction following segmentation, yet generalization remains limited. The key bottleneck is the high cost of curating diverse, high-quality pixel annotations paired with rich linguistic supervision leading to brittle performance under distribution shift.
Therefore, we present \textbf{CORA}, a semi-supervised reasoning segmentation framework that jointly learns from limited labeled data and a large corpus of unlabeled images. CORA introduces three main components: 1) conditional visual instructions that encode spatial and contextual relationships between objects; 2) a noisy pseudo-label filter based on the consistency of Multimodal LLM's outputs across semantically equivalent queries; and 3) a token-level contrastive alignment between labeled and pseudo-labeled samples to enhance feature consistency. These components enable CORA to perform robust reasoning segmentation with minimal supervision, outperforming existing baselines under constrained annotation settings. CORA achieves state-of-the-art results, requiring as few as 100 labeled images on Cityscapes, a benchmark dataset for urban scene understanding, surpassing the baseline by $+2.3\%$. Similarly, CORA improves performance by $+2.4\%$ with only 180 labeled images on PanNuke, a histopathology dataset.
\end{abstract}
    
\section{Introduction}
\label{sec:intro}
\begin{figure}[h!]
\centering
\newcommand{\MethodNameBJ}{\textbf{MethodNameBJ}}

\includegraphics[width=1.0\linewidth]{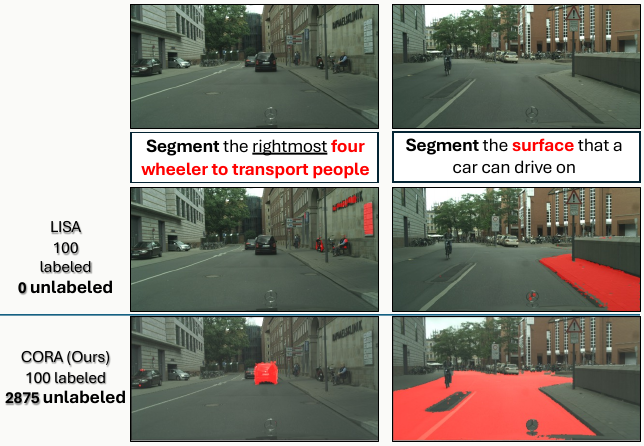}
     \caption{\textbf{Training reasoning segmentation systems with semi-supervised semantic segmentation supervision.} The second row shows results from LISA \cite{lai2024lisa} trained on 100 labeled images, while the third row shows results from CORA (Ours) on 100 labeled images and \textbf{2,875 unlabeled images}. We note that unlike LISA \cite{lai2024lisa}, which is limited to fully supervised settings, our method effectively leverages 2,875 unlabeled images alongside just 100 labeled examples, demonstrating that unlabeled data can significantly enhance reasoning-based segmentation performance.
     }
\label{fig: teaser}
\end{figure}

In complex scenarios like autonomous driving, mere segmentation is not enough. Systems must also reason about what they perceive. For example, identifying a pedestrian is one step, but reasoning about their intention, whether they are about to cross or hesitate, is an important safety consideration. Similarly, in medical imaging, the detection of a tumor is useful, but determining whether it is malignant and localizing its affected surrounding tissues is essential for diagnosis and treatment.  Reasoning segmentation \cite{lai2024lisa, wang2024llm, wang2025segllm}, enabled by Multimodal Large Language Models (MLLMs), bridges the gap between visual perception and human-like reasoning, facilitating more reliable decision-making in these complex environments.

However, existing MLLMs remain largely confined to COCO-like datasets \cite{kazemzadeh2014referitgame, yu2016modeling, mao2016generation} with limited generalization beyond generic object recognition tasks. A major bottleneck is the lack of diverse, annotated data\cite{JingyiICCV21,Xu2022GeneratingRS,Xu2023FSOD} for reasoning about object intents, particularly in domains that require expert knowledge. In autonomous driving, understanding pedestrian behavior not only requires pixel-level annotations but also expert-designed linguistic queries that capture subtle intent (e.g., ``\textit{segment the pedestrian about to cross}"). Similarly, in medical imaging, both delineating tumor boundaries and formulating descriptive queries (e.g., ``\textit{malignant mass near the liver}") demand input from trained radiologists. Generating such high-quality visual and linguistic supervision is labor-intensive and expensive. This limits model generalization and hinders the real-world deployment of reasoning-based segmentation systems.

Thus, we explore the feasibility of training reasoning segmentation models with only weak supervision by formulating the task in a semi-supervised setting. In this setup, a small number of labeled masks are available alongside abundant unlabeled images. At first glance, this resembles semi-supervised semantic segmentation~\cite{di2021semi, hu2021semi, li2020semi,Xu2023ZeroShotOC,Le_2020_ECCV,le2020physicsbased,Le_RS22}, which has been extensively studied. However, extending it to reasoning segmentation is far from trivial. Traditional semi-supervised segmentation operates in a closed-vocabulary regime, where each mask is tied to a fixed class label. Reasoning segmentation, by contrast, must ground free-form linguistic queries in open-vocabulary concepts and contextual relations. Moreover, with only semantic masks available, it is unclear how to generate the kind of query–segment supervision reasoning segmentation requires. These differences imply that existing semi-supervised techniques cannot be directly applied without rethinking both the supervision and the training process.

To address these challenges, we propose \textbf{CORA} (\textbf{\underline{C}}onsistency-guided \underline{\textbf{O}}bject \underline{\textbf{R}}elational \underline{\textbf{A}}lignment), a semi-supervised framework that advances reasoning segmentation by addressing both the scarcity of labeled data and the underutilization of abundant unlabeled data. CORA is motivated by two core principles: in low-label regimes, supervision must be enriched beyond raw annotations, and in high-unlabeled regimes, only reliable signals should be extracted. More specifically, we incorporate three complementary components into CORA:
\begin{enumerate}
    \item \textbf{Conditional visual instructions.} We generate diverse, spatially grounded queries from segmentation masks, enriching supervision per labeled image and enabling the model to capture object interactions and contextual reasoning under low-label regimes.
    \item \textbf{Output consistency filtering.} To reduce noise in unlabeled supervision, we assess the stability of predictions across semantically equivalent query rephrasings and retain only consistent pseudo-labels.

    \item \textbf{Token-level feature alignment.} We introduce a contrastive loss that enforces semantic consistency between labeled and pseudo-labeled objects of the same class, improving representation robustness.
\end{enumerate}

Together, these components establish the first framework for semi-supervised reasoning segmentation, enabling robust learning under minimal supervision and providing a practical guideline for converting plain semantic masks into pseudo reasoning supervision using off-the-shelf models, bridging the gap between conventional semantic segmentation data and reasoning-based training. 

To summarize, our contributions are:
\begin{enumerate}
    \item We formulate the task of semi-supervised reasoning segmentation, making reasoning-based perception feasible in domains with scarce annotations.  
    \item We propose \textbf{CORA}, a framework that enriches limited supervision with conditional visual instructions and exploits unlabeled data through output consistency filtering and token-level contrastive alignment.  
    \item We demonstrate that CORA achieves new state-of-the-art performance on both \textbf{Cityscapes} and \textbf{PanNuke}.  
\end{enumerate}


\section{Related Works}
\subsection{Semi-supervised segmentation} 
Early semi-supervised segmentation used GANs to align predictions with ground-truth distributions \cite{goodfellow2014generative, mittal2019semi, souly2017semi}, while self-training methods generated pseudo-labels for iterative retraining \cite{grandvalet2004semi, lee2013pseudo}. To reduce error drift, strategies such as confidence weighting \cite{feng2022dmt, yang2022st++}, curricula \cite{ma2023enhanced, xu2021dash}, contrastive learning \cite{wang2022semi}, and symbolic reasoning \cite{liang2023logic} were introduced. Pseudo-labeling remains the popular approach \cite{yang2023revisiting, hu2021semi, howlader2024weighting}, typically within a teacher–student framework. However, these methods are image-only and closed-set, and they do not integrate with language or reasoning. In CORA, pseudo-labels not only supervise mask generation but also serve as class priors for constructing text instructions.
\subsection{Reasoning Segmentation Methods} 
Referring segmentation methods \cite{hu2016segmentation,yang2022lavt,zou2023segment} segment objects from text descriptions but lack reasoning beyond direct references, making them weak baselines for the more challenging reasoning segmentation task. To address this, MLLMs have been integrated into segmentation frameworks. LISA \cite{lai2024lisa} combines LLaVA \cite{liu2023visual} with SAM \cite{kirillov2023segment}, but fine-tuning SAM’s mask decoder often leads to coarse boundaries. Recent works such as LLM-Seg \cite{wang2024llm} and SegLLM \cite{wang2024segllm} incorporate spatial and contextual cues from text, yet rely on full supervision, limiting their applicability in low-label regimes.

Building on these works, we propose CORA, a semi-supervised framework that improves reasoning segmentation through conditional visual prompts, output consistency–based pseudo-label filtering, and token-level contrastive alignment, enabling robust performance with limited annotations.

\subsection{Handling Noisy Pseudo Labels } 
Uncertainty has been widely explored in semantic segmentation \cite{kendall2015bayesian, sedai2019uncertainty} and semi-supervised learning \cite{yu2019uncertainty, pham2021meta} to filter noisy pseudo-labels\cite{durasov2024enabling,durasov2024zigzag,Le_CVPRW19,Xu2024ASQ,JingyiICCV21} and improve reliability.
Common strategies include Monte Carlo (MC) Dropout \cite{gal2016dropout} for epistemic uncertainty and ensemble methods \cite{lakshminarayanan2017simple} for estimation.
Recent approaches such as UniMatch \cite{yang2023revisiting} and UCC \cite{fan2022ucc} leverage uncertainty calibration to refine pseudo-labels.

In contrast, uncertainty estimation in MLLMs is still emerging.
Although LLM-based approaches \cite{wang2024llm, yang2023revisiting} show promise for reasoning-based segmentation, they often lack quantifiable uncertainty.
Vision-language works \cite{upadhyay2023probvlm, chandu2024certainly} attempt to address this but overlook semi-supervised reasoning segmentation.
In this work, we exploit the reasoning ability of MLLMs by measuring consistency across semantically equivalent query rephrasings, yielding per-pixel uncertainty estimates that enable more effective pseudo-label filtering.
\section{Proposed Method}

 \begin{figure*}[ht!]
 \centering
 \includegraphics[width=\linewidth]{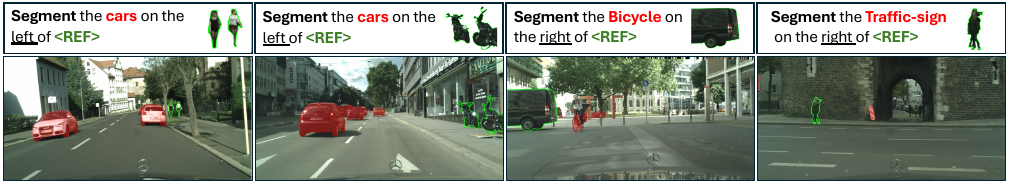}
 \caption{\textbf{Illustration of Conditional-relationship Visual Instruction Set used for training:} \textcolor{red}{Target object} segmentation conditioned on its contextual relationship with the \textcolor{green}{reference object (Anchor)}.
      }
 \label{fig:visual_instruction_set}
 \end{figure*}

In this section, we detail the working principle of our proposed Consistency-guided Object Relational Alignment (CORA) which is a novel semi-supervised framework for training reasoning-based segmentation models. The objective with CORA is to make reasoning segmentation feasible when labeled data is scarce but large amounts of unlabeled data are available. CORA is guided by two principles: 1) enrich supervision when only limited labels exist, and 2) extract only reliable signals when leveraging unlabeled images.

To achieve this, our framework combines three components:
\begin{enumerate}
    \item \textit{Conditional visual instructions} expand the supervision provided by labeled masks by turning them into diverse, spatially grounded queries that describe relationships between objects.
    \item \textit{Output consistency filtering} improves the reliability of pseudo-labels on unlabeled data by retaining only predictions that remain stable across query rephrasings.
    \item \textit{Token-level contrastive alignment} enforces feature consistency between labeled and pseudo-labeled objects, ensuring robust class representations.
\end{enumerate}

We train CORA in three stages. First, we pretrain the model on semantic and attribute-based instructions to learn object categories. Second, we introduce conditional visual instructions to capture spatial and contextual relationships. Finally, we incorporate unlabeled images, refining them with consistency filtering and token-level alignment. In the following subsections, we describe each component of CORA in detail. We begin with conditional visual instructions, which strengthen supervision in labeled data. We then turn to output consistency filtering and token-level contrastive alignment, which together make it possible to exploit unlabeled data effectively. Finally, we present the overall training objective and describe how we generate the instruction set used to train CORA.

\subsection{Conditional Visual Instructions}
Following LISA~\cite{lai2024lisa} and LLM-SEG~\cite{wang2024llm}, we first generate semantic and attribute-based instructions from the limited labeled data (Section~\ref{sec:instructionset}). However, with as few as 99 labeled images in the $\tfrac{1}{30}$ split of Cityscapes, the number of instruction–image pairs is too small to provide strong supervision. Prior work~\cite{zhang2024visually} has shown that increasing the presence of each class during multimodal LLM training improves performance on those classes. Motivated by this, we introduce conditional visual instructions to enrich supervision under low-label conditions.

As illustrated in Fig.~\ref{fig:visual_instruction_set}, we randomly select an anchor object from the image (e.g., “the bike”) and use it to condition the segmentation of a target object (e.g., “the cars”). The spatial relation between anchor and target (e.g., left, right, above) is embedded into the query. This design has two benefits: it multiplies the number of available training instructions per class, and it encourages the model to learn spatial context, which is essential for reasoning-based segmentation.

Formally, given a labeled image $x^l_{img}$, its segmentation mask $M^l$, and a conditional instruction $x^l_{vp}$, the multimodal LLM $\mF_{vlm}$ generates a textual response:
\begin{equation}\label{eq1}
    \hat{y}_{txt} = \mF_{vlm}(x_{img}^l, x^l_{vp})
\end{equation}

Following LISA \cite{lai2024lisa}, the multimodal LLM response $\hat{y}_{txt}$ includes one \texttt{<SSEG\textsuperscript{$l$}>}  token for the binary mask to be generated. Simultaneously, the input image $x_{img}^l$ is fed through an image encoder $\mF_{enc}$ to generate the embeddings $\mE_{img}^{l}$. Finally, the embedding corresponding to the \texttt{<SSEG\textsuperscript{$l$}>} token and $\mE_{img}^{l}$ are fed to the decoder $\mF_{dec}$, yielding the final segmentation mask $\hat{M}^l$ as:
\begin{equation}\label{eq2}
    \mE_{img}^{l} = \mF_{enc}(x^l_{img}),
\end{equation}
\begin{equation}\label{eq3}
    \hat{M}^l = \mF_{dec}(\mE_{img}^{l}, \texttt{<SSEG\textsuperscript{$l$}>})
\end{equation}

The training loss on labeled images is the sum of a segmentation loss and a language modeling loss:
\begin{equation}\label{eq4}
    \mL^l_{seg} = \text{BCE}(\hat{M}^l,M^l) +  \text{DICE}(\hat{M}^l,M^l)
\end{equation}
\begin{equation}\label{eq5}
    \mL^l = \alpha \mL_{ce} + \mL^l_{seg}
\end{equation}
where $\mL_{ce}$ is the cross-entropy loss between the generated and ground-truth text, and $\alpha$ balances the two terms.

\subsection{Leveraging Unlabeled Data}
Unlabeled data is vital for semi-supervised segmentation, but standard confidence-thresholding~\cite{yang2023revisiting, hu2021semi, zhao2023augmentation} is brittle: high-confidence predictions can still be wrong~\cite{rizve2021defense}, while low-confidence ones may include valuable hard examples~\cite{ma2023enhanced, howlader2024weighting}. The challenge is even harder in reasoning segmentation, where outputs depend on both the image and the textual query.

We address this with two consistency-based strategies: 1. Output consistency–driven pseudo-labeling estimates per-pixel reliability by measuring prediction variance across semantically equivalent queries, down-weighting unstable regions. 2. Token-level feature alignment, inspired by GLUS~\cite{lin2025glus}, enforces that \texttt{<SSEG>} tokens for the same class remain consistent across labeled and unlabeled data using a SimCLR-style contrastive loss~\cite{chen2020simple}.

 \label{sec:approach}
 \begin{figure*}[ht!]
 \centering
 \includegraphics[width=\linewidth]{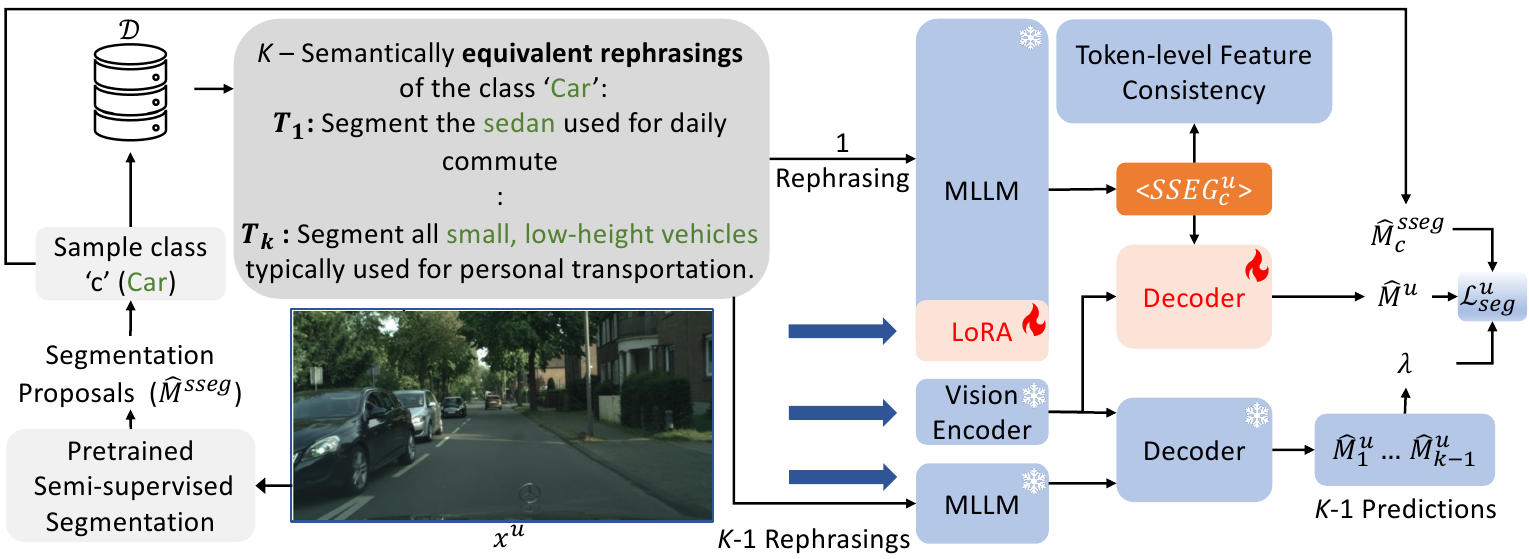}
 \caption{\textbf{Framework of our approach leveraging unlabeled images for reasoning segmentation } CORA is trained on unlabeled images using pseudo-labels from a pretrained semi-supervised segmentation model, with output consistency from a multi-modal LLM used to reduce pseudo-label noise.
      }
 \label{fig: overallmodel1}
 \end{figure*}

\begin{figure}[h!]
\centering
\includegraphics[width=0.8\linewidth]{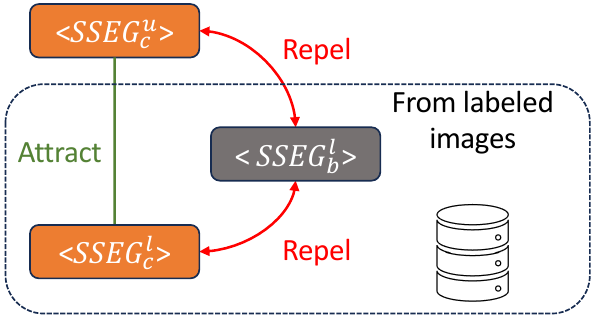}
     \caption{\textbf{Token-level Feature Consistency Alignment.} Minimize the distance between the unlabeled token feature (\texttt{<SSEG\rlap{\textsuperscript{$u$}}\textsubscript{$c$}>}) and the same-class labeled token (\texttt{<SSEG\rlap{\textsuperscript{$l$}}\textsubscript{$c$}>}), while maximizing its distance from different-class labeled tokens (\texttt{<SSEG\rlap{\textsuperscript{$l$}}\textsubscript{$b$}>})
     }
\label{fig: contrastive}
\end{figure}

\subsubsection{Output consistency-driven pseudo-labeling} 
Pseudo-labels from semi-supervised segmentation models are often noisy, and confidence-thresholding heuristics are unreliable. Instead, we exploit the reasoning ability of multimodal LLMs by measuring how consistent predictions remain across semantically equivalent query rephrasings \cite{khan2024consistency}. This consistency provides a per-pixel estimate of uncertainty, allowing us to filter noisy labels more effectively.

More specifically, we introduce an adaptive per-pixel learning weight where the weight is quantified via output variance across equivalent queries. Given an unlabeled image $x^u$, we run an off-the-shelf semi-supervised segmentation model to obtain its per-pixel label prediction. For a sampled class $c$, we extract the binary mask $\hat{M}^{sseg}$ from this prediction and retrieve $k$ precomputed queries for class $c$ from $\mathcal{D}$. One query is used to train the multimodal LLM, producing prediction $\hat{M}^u$. The remaining $k-1$ queries are passed through a frozen LLM and decoder to compute per-pixel variance, which defines an adaptive weight $\lambda_i$:

\begin{equation}\label{eq6}
    \lambda_i = \frac{1}{k-1}\sum_{j=1}^{k-1}(\hat{M}_{ij}^{u} - \frac{1}{k-1}\sum_{l=1}^{k-1} \hat{M}_{il}^{u})^2
\end{equation}

Here, $\hat{M}_{ij}^{u}$ and $\hat{M}_{il}^{u}$ are predictions from different query rephrasings. Unstable pixels receive lower weights, while stable ones are emphasized.

The final loss on unlabeled data combines binary cross-entropy and Dice terms, modulated by $\lambda_i$:

\begin{equation}\label{eq7}
    \mL^u_{seg} =  \sum_{i=1}^{WH}\lambda_i (\text{BCE}(\hat{M}_{i}^u ,\hat{M}_{i}^{sseg}) +  \text{DICE}(\hat{M}_{i}^u,\hat{M}_{i}^{sseg})),
\end{equation}
where $W$ and $H$ are the image dimensions, and $\hat{M}^{sseg}$ is the pseudo-mask from the off-the-shelf semi-supervised segmentation model.

\subsubsection{Token-Level Feature Consistency Alignment}

Output-level consistency reduces noise in pseudo-labels, but it does not guarantee that features learned from labeled and unlabeled data remain semantically aligned. To address this, we introduce a token-level consistency loss that enforces alignment at the object representation level.

Our approach is inspired by GLUS~\cite{lin2025glus}, which showed that object tokens (\texttt{<SSEG>}) from different referring expressions of the same object should remain consistent across video frames.
We extend this idea to the semi-supervised image setting. Here, we enforce that the \texttt{<SSEG>} token—the output token from the multimodal LLM corresponding to a given target class—remains consistent across labeled and unlabeled data (Fig.~\ref{fig: contrastive}). We achieve this by minimizing the distance between \texttt{<SSEG>} tokens of same class objects from these two data sources. We achieve this by maintaining a token bank \cite{el2023learning,he2020momentum,lin2025glus,he2024decoupling} of different tokens from labeled images.

Formally, the object-level contrastive loss follows SimCLR~\cite{chen2020simple}, where the feature vector $v$ of a \texttt{<SSEG>} token from unlabeled images is supervised using a positive sample $k^+$ and a set of negative samples $k^-$ both sampled from labeled images stored in the token bank. The loss is defined as:

\begin{equation}
\mathcal{L}^u_t = -\sum_{v,k^+} \log \frac{\exp(\text{sim}(v, k^+)/\tau)}{\exp(\text{sim}(v, k^+)/\tau) + \sum\limits_{k^-} \exp(\text{sim}(v, k^-)/\tau)}
\end{equation}
where $\text{sim}(\cdot, \cdot)$ denotes cosine similarity, $\tau$ is a temperature scaling parameter, $k^+$ denotes  \texttt{<SSEG>} tokens from objects of the same class as $v$,  and $k^-$ denotes  \texttt{<SSEG>} tokens from objects of other classes.

This token-level alignment enforces that class-specific representations remain semantically consistent across labeled and unlabeled samples, improving generalization in reasoning-based segmentation.


 \begin{figure*}[ht!]
 \centering
 \includegraphics[width=\linewidth]{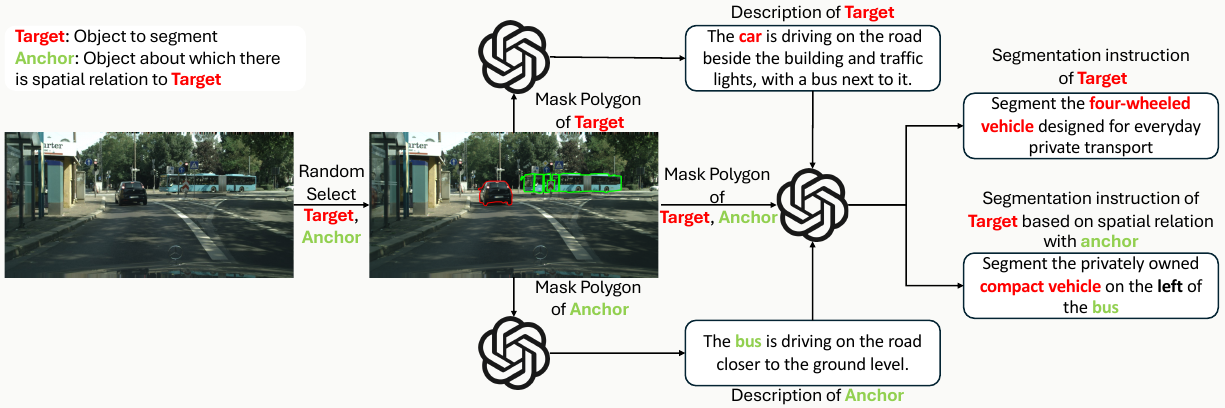}
 \caption{\textbf{Pipeline for generating our dataset for training CORA} Given an image, two random objects are selected as target and anchor from the segmentation mask. Descriptions of each are generated using the image and their mask polygons. These textual descriptions, together with the polygons, are then used to create the segmentation instruction set. The system prompt to generate the conditional visual instruction is in Supplementary (Section 7.2)
      }
 \label{fig: data_pipeline}
 \end{figure*}
\subsection{Training}
CORA training follows a three-stage framework, each stage running for an equal number of iterations. We first train CORA following LISA \cite{lai2024lisa} and SegLLM \cite{wang2024segllm} on  semantic segmentation and attribute-based instruction set, enabling the model to learn individual object classes. Next, we train CORA on spatial conditional visual instruction set using labeled images, encouraging the model to learn spatial relationships and contextual dependencies within a scene. In the final stage, we incorporate unlabeled images, refining the model using pseudo-labels from a semi-supervised segmentation network.

During Stage 2 training on the spatial conditional visual instructions, we utilize an anchor image mask and its corresponding bounding box, encoded separately using an anchor image encoder and a bounding box encoder as proposed by Seg-LLM \cite{wang2024segllm}.

The overall loss function to train CORA on both labeled and unlabeled image can be written as:
\begin{equation}\label{eq10}
    \mL = \mL^l + \sigma (\mL^u_{sseg} + \mL^u_t )
\end{equation}
Where $\sigma$ controls the contribution of loss from unlabeled images.

\subsection{Instruction Set for CORA}\label{sec:instructionset}
\subsubsection{Data Sources}
We construct reasoning segmentation dataset based on two widely used datasets in the field of autonomous driving and histopathology: Cityscapes \cite{Cordts2016Cityscapes} and PanNuke dataset \cite{gamper2019pannuke}. We use the segmentation annotations from these datasets to generate visual instruction set, the details of this approach is detailed in subsequent sections. The overall pipeline can be seen in Fig.~\ref{fig: data_pipeline}.
\noindent
\subsubsection{Visual Instruction Set }
We design various instruction sets from labeled images, tailored to the different types of relationships:

\begin{itemize}
    \item \textbf{Conditional-relationship Visual Instruction Set}: Illustrated in Fig.~\ref{fig:visual_instruction_set}, these visual instructions require the model to segment a target object based on its spatial relationship with an anchor object in the image. An example query is: \textit{`` Can you segment the $<$target$>$ that is $<$spatial relationship$>$ to the $<$masked image of anchor$>$?"}. To create such instructions (Fig.~\ref{fig: data_pipeline}), we use GPT-4o with a set of designed prompts.  First, we use GPT-4o to generates textual descriptions of the target and anchor objects.  Then, given the image, their mask polygons \cite{ranasinghe2024learning}, and these descriptions, GPT-4o produces question–answer pairs that encode the spatial relation between the two objects.  To reduce ambiguity, we specify the camera as the frame of reference \cite{zhang2024vision} for the spatial relation. The full prompts used for generating these conditional instructions is included in the supplementary (Sec.~7.2).
    \item \textbf{Attribute-based Instruction Set}: These instructions require the model to segment objects based on their attributes rather than class names. For instance, to segment cars, a query could be: \textit{`` Can you segment the four wheeled vehicle used for personal transport?"}. To generate these instructions, we employ GPT-4o with visual descriptions of the target object. To ensure attribute-level uniqueness and disambiguate between visually similar objects (e.g., \textit{car} vs. \textit{bus}), we provide both the image and mask polygons of all objects in the scene. The final prompt to generate the attribute-based instruction set is in the supplementary (Sec. 7.3)
    \item  \textbf{Semantic Segmentation Instruction Set}: We construct these instructions by fitting class labels into segmentation query templates.
\end{itemize}
For unlabeled images, we maintain a database of precomputed queries per class, generated by incorporating distinguishing attributes into segmentation query templates. To ensure query diversity and natural language variability, we utilize the web-based GPT-4o to generate multiple templates per class, enabling more robust and linguistically varied segmentation instructions.

\smallskip
\noindent
\textbf{Dataset Verification.} To assess the quality of our LLM-generated instructions, we conducted a human evaluation with 14 participants on 50 randomly sampled images. Each query–image pair was rated on a 1–5 correctness scale, yielding an average score of 3.8. This confirms that combining mask contours with visual descriptions produces high-quality instruction sets.

\section{Experiments}

\subsection{Implementation Details}
\textbf{Model structure and training settings.} We use CLIP-ViT-Large \cite{radford2021learning} as the image encoder, LLaVA-v1.5-7B \cite{liu2023visual} as the base multimodal LLM, and follow LISA \cite{lai2024lisa} in incorporating SAM ViT-H \cite{kirillov2023segment} as the vision backbone. Pseudo-labels for unlabeled images are generated using UniMatch \cite{yang2023revisiting} and SemiVL \cite{hoyer2024semivl}, both pretrained on the same partitions as CORA without requiring extra annotations. Training is performed on NVIDIA A6000 GPUs with batch size 8 (2 per device), AdamW optimizer, and learning rate $2e^{-5}$. The text generation loss weight $\alpha$ and unlabeled loss weight $\sigma$ are set to 1.0 and 0.001, respectively. For consistency-driven pseudo-labeling, we sample $K{=}7$ queries per class to compute per-pixel weights from prediction variance.

\smallskip
\noindent
\textbf{Dataset.} We conduct qualitative and quantitative comparisons against state-of-the-art (SOTA) models on reasoning segmentation benchmarks using Cityscapes and PanNuke datasets. \textbf{Cityscapes dataset}\cite{Cordts2016Cityscapes} is designed for urban scene understanding,
consisting of $30$ classes, of which only $19$ classes are used for scene parsing evaluation. It has $2975$, $500$, and $1525$ images for training, validation, and testing,
respectively.  \textbf{PanNuke dataset}\cite{gamper2019pannuke} is a histopathology dataset with $7,904$ images, covering $19$ tissue types, split into $5381$ training and $2523$ validation images.

\smallskip
\noindent
\textbf{Evaluation Metrics.} Following LISA \cite{lai2024lisa} and SegLLM \cite{wang2024segllm}, we use cumulative IoU (cIoU) as our primary metric.

\begin{table}[]
\begin{subtable}{\linewidth}
\centering
\resizebox{\linewidth}{!}{
\begin{tabular}{@{}l|cccc|cccc@{}}
           & \multicolumn{4}{c|}{UniMatch\cite{yang2023revisiting} (unlabeled)}                     & \multicolumn{4}{c}{SemiVL\cite{hoyer2024semivl} (unlabeled)}                       \\ \cmidrule(l){2-9} 
\rowcolor{violet!20} Methods    & 1/30          & 1/16          & 1/8           & 1/4           & 1/30          & 1/16          & 1/8           & 1/4           \\ \midrule
    \textit{Supervised} (only labeled)        & 48.3          & 50.7          & 53.1          & 55.2          & 48.3          & 50.7          & 53.1          & 55.2          \\
VLT\cite{ding2021vision}        & 49.9          & 52.2          & 54.8          & 55.7          & 51.2          & 53.1          & 54.9          & 56.0          \\
LAVT\cite{yang2022lavt}       & 51.8          & 53.9          & 55.6          & 57.0          & 53.5          & 55.4          & 56.7          & 57.7          \\
SEEM\cite{zou2023segment}       & 51.3          & 54.2          & 56.2          & 58.6          & 55.9          & 58.4          & 62.4          & 64.1          \\
LISA-7B\cite{lai2024lisa}    & 52.7          & 55.8          & 58.4          & 61.1          & 58.2          & 61.3          & 63.3          & 65.4          \\
NexT-Chat\cite{zhang2023next}  & 52.9          & 55.7          & 57.8          & 59.4          & 58.7          & 61.2          & 63.0          & 64.9          \\ 
SegLLM\cite{wang2024segllm}(single-round)     & 54.2          & 56.9          & 59.2          & 61.2          & 60.0          & 63.1          & 65.7          & 67.4          \\ \midrule
CORA (Ours) & \textbf{56.9} & \textbf{58.2} & \textbf{60.5} & \textbf{62.3} & \textbf{62.2} & \textbf{65.0} & \textbf{66.9} & \textbf{68.1} \\ \bottomrule
\end{tabular}}
\end{subtable}
\caption{\textbf{Quantitative comparison of SOTA methods on the Cityscapes dataset}. We report cumulative IoU (cIoU) under different partition protocols. CORA consistently outperforms all competing methods.}
\label{tab:cityscapes}
\end{table}

\begin{table}[]
\begin{subtable}{\linewidth}
\centering
\resizebox{\linewidth}{!}{
\begin{tabular}{@{}l|cccc|cccc@{}}
           & \multicolumn{4}{c|}{UniMatch\cite{yang2023revisiting} (unlabeled)}                     & \multicolumn{4}{c}{SemiVL\cite{hoyer2024semivl} (unlabeled)}                       \\ \cmidrule(l){2-9} 
\rowcolor{violet!20} Methods    & 1/30          & 1/16          & 1/8           & 1/4           & 1/30          & 1/16          & 1/8           & 1/4           \\ \midrule
\textit{Supervised} (only labeled)       & 38.0          & 41.5          & 44.7          & 48.0          & 38.0          & 41.5          & 44.7          & 48.0           \\
VLT\cite{ding2021vision}        & 40.4          & 43.8          & 46.5             & 49.2             & 43.4          & 47.1          & 50.4             & 53.4             \\
LAVT\cite{yang2022lavt}       & 41.7          & 44.9          & 47.5             & 50.2             & 45.1          & 48.5          & 51.5             & 53.9             \\
SEEM\cite{zou2023segment}       & 43.2          & 46.1          & 48.6          & 50.6          & 47.0          & 50.2          & 52.9          & 55.0          \\
LISA-7B\cite{lai2024lisa}    & 45.1          & 47.6          & 49.6          & 51.1          & 48.7          & 51.6          & 54.1          & 56.0          \\
NexT-Chat\cite{zhang2023next}  & 46.8          & 48.9          & 50.7          & 52.0          & 50.6          & 53.0          & 55.1          & 56.5          \\
SegLLM\cite{wang2024segllm} (single-round)     & 47.2          & 49.3          & 51.1          & 52.3          & 51.2          & 53.7          & 55.8          & 57.3          \\ \midrule
CORA (Ours) & \textbf{49.6} & \textbf{51.4} & \textbf{52.3} & \textbf{53.5} & \textbf{53.9} & \textbf{55.8} & \textbf{57.2} & \textbf{58.5} \\ \bottomrule
\end{tabular}}
\end{subtable}
\caption{\textbf{Quantitative comparison of SOTA methods on the PanNuke dataset}. We report cumulative IoU (cIoU) under different partition protocols. CORA consistently outperforms all competing methods.}
\label{tab:pannuke}
\end{table}

\begin{table}[]
\centering
\begin{subtable}{\linewidth}
\resizebox{\linewidth}{!}{%
\begin{tabular}{@{}l|cc|cc|cc|cc@{}}
        & \multicolumn{2}{c|}{1/30} & \multicolumn{2}{c|}{1/16} & \multicolumn{2}{c|}{1/8} & \multicolumn{2}{c}{1/4} \\ \cmidrule(l){2-9} 
\rowcolor{violet!20} Methods & Test      & Test (LQ)     & Test      & Test (LQ)     & Test     & Test (LQ)     & Test     & Test (LQ)    \\ \midrule
LISA \cite{lai2024lisa}    & 40.8         & 50.0             & 42.5         & 42.9             & 44.1        & 44.2             & 47.3        & 47.5            \\
CORA (Ours)   & \textbf{42.3}         & \textbf{51.3}             &\textbf{43.7}          & \textbf{43.9}             & \textbf{45.0}        & \textbf{45.3}             & \textbf{48.0}        & \textbf{48.5}            \\ \bottomrule
\end{tabular}}
\end{subtable}
\caption{\textbf{Quantitative comparison on the ReasonSeg dataset.} We train both LISA and CORA on different partition protocols of the Cityscapes dataset and evaluate them on the ReasonSeg dataset \cite{lai2024lisa}, reporting cumulative IoU (cIoU) for each protocol. Here, LQ denotes long query. We use UniMatch \cite{yang2023revisiting} as the off-the-shelf semi-supervised method for this experiment.}
\label{tab:lisa}
\end{table}

 \begin{figure*}[ht!]
 \centering
 \includegraphics[width=\linewidth]{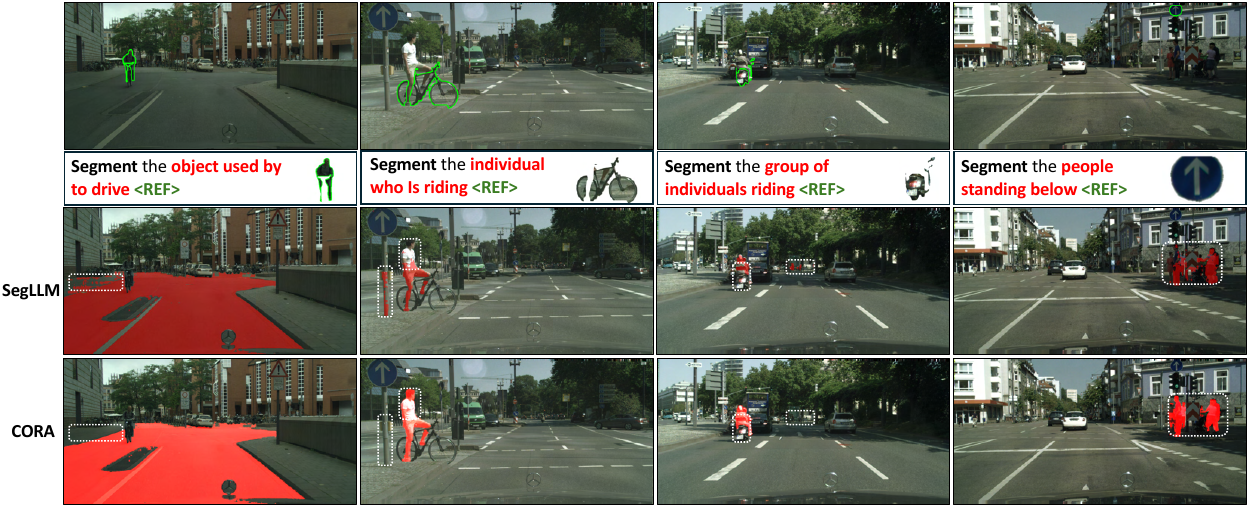}
 \caption{\textbf{Qualitative Results on Cityscapes dataset: } Segmentations of SegLLM \cite{wang2024segllm} and CORA are shown. White boxes highlight regions where CORA produces more accurate segmentations. The results demonstrate that CORA is better able to segment the \textcolor{red}{target object} based on its relationship to the \textcolor{ForestGreen}{reference object} described in the query.
      }
 \label{fig: qualitative results}
 \end{figure*}
\subsection{Quantitative Results}
We evaluate CORA on Cityscapes and PanNuke under semi-supervised settings with four splits ($\tfrac{1}{30}$, $\tfrac{1}{16}$, $\tfrac{1}{8}$, $\tfrac{1}{4}$) following SemiVL \cite{hoyer2024semivl}, where only a fraction of training data is labeled (e.g., $\tfrac{1}{30}$). We also train CORA and LISA \cite{lai2024lisa} on multiple Cityscapes partitions and evaluate them on the reasoning segmentation benchmark ReasonSeg \cite{lai2024lisa}. SegLLM is adapted to single-round mode, and baseline details with LISA prompts are in the supplementary. CORA consistently outperforms state-of-the-art baselines across all splits, highlighting its effectiveness for semi-supervised reasoning segmentation.

\smallskip
\noindent
\textbf{Results on Cityscapes Dataset.} Table \ref{tab:cityscapes} compares CORA with SOTA baselines. CORA outperforms all methods across partitions using both UniMatch and SemiVL, with the largest gains on the $\frac{1}{30}$ split (\textcolor{teal}{$+2.7\%$} with UniMatch, \textcolor{teal}{$+2.2\%$} with SemiVL).

\smallskip
\noindent
\textbf{Results on PanNuke Dataset.} Table \ref{tab:pannuke} shows that CORA outperforms SOTA baselines across all partitions, with the largest gain on the $\frac{1}{30}$ split (\textcolor{teal}{$2.4\%$} with UniMatch, \textcolor{teal}{$2.7\%$} with SemiVL).

\smallskip
\noindent
\textbf{Results on ReasonSeg Dataset.} We train CORA and LISA \cite{lai2024lisa} on multiple Cityscapes partitions and evaluate on the existing reasoning segmentation dataset ReasonSeg \cite{lai2024lisa}. As shown in Table \ref{tab:lisa}, CORA achieves the largest gain on the $\tfrac{1}{30}$ split, surpassing LISA by \textcolor{teal}{1.5\%}. We use UniMatch \cite{yang2023revisiting} as the off-the-shelf semi-supervised model.

These results demonstrate the effectiveness of CORA in both urban scene understanding and biomedical reasoning segmentation tasks, consistently surpassing SOTA baselines across all data partitions.
When trained on different data partitions of the Cityscapes dataset, CORA also achieves higher cumulative IoU (cIoU) on the widely used ReasonSeg dataset across all partitions.
Notably, CORA attains the largest improvement in the lowest-labeled data regime ($\frac{1}{30}$), highlighting its effectiveness under limited supervision.
Moreover, its consistently superior performance across various off-the-shelf semi-supervised segmentation models demonstrates CORA’s robustness in integrating pseudo-labels from such models.
\subsection{Qualitative Results}
Fig.~\ref{fig: qualitative results} compares CORA with SegLLM \cite{wang2024segllm}. CORA yields more accurate segmentations by using conditional visual instructions, consistency-based pseudo-label refinement, and token-level feature alignment.
\subsection{Limitations} 
\begin{figure}[ht!]
\centering

\includegraphics[width=1.0\linewidth]{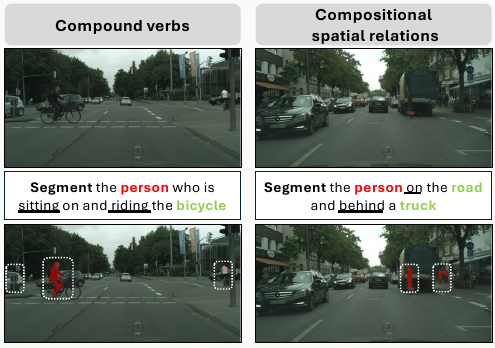}
     \caption{\textbf{Limitations.} We observe that CORA makes segmentation errors on complex queries, particularly those involving compound verbs and compositional spatial relations.
     }
\label{fig: limitations}
\end{figure}

In Fig.~\ref{fig: limitations}, we illustrate the limitations of CORA. It makes segmentation errors on complex queries, particularly those involving compound verbs and compositional spatial relations. Even though CORA explicitly trains the model to capture relations between target and anchor objects in an image, handling composite spatial relations and compound verbs requires incorporating more complex training queries. This highlights an important direction for future research.
\section{Ablation Study}
We conduct experiments to study the impact of individual components in our approach. UniMatch is used as the off-the-shelf semi-supervised method for generating pseudo-labels on unlabeled images.


\begin{table}[ht!]
\centering
\begin{tabular}{@{}ccc|ccc@{}}
\toprule
CVI & OCPL & TFCA & 1/30 & 1/16 & 1/8  \\ \midrule
   &     &     & 52.9 & 55.8 & 58.5 \\
\checkmark   &      &     & 55.1 & 57.3 & 59.5 \\
   & \checkmark    &     & 54.4 & 56.9 & 59.0 \\
   &     & \checkmark    &   53.5   &  56.2    &  59.0    \\
\checkmark   & \checkmark    &     & 56.3     & 57.9     &  59.9    \\
\checkmark   &     & \checkmark    &  55.8    &   57.5   &    59.6  \\
\checkmark   & \checkmark    & \checkmark    & \textbf{56.9}     & \textbf{58.2}     &    \textbf{60.5}  \\ \bottomrule
\end{tabular}
\caption{\textbf{CORA Ablation}: CVI (Conditional Visual Instructions), OCPL (Output Consistency–driven Pseudo-labeling), and TFCA (Token-level Feature Consistency Alignment).}
\label{tab:Ablationcomponent}
\end{table}

\smallskip
\noindent
\textbf{Analysis of different components in CORA.} We ablate each component of CORA in Table~\ref{tab:Ablationcomponent}. As a baseline, we follow LISA~\cite{lai2024lisa} with attribute-based instructions for labeled data and confidence-thresholded pseudo-labels for unlabeled data. Adding Conditional Visual Instructions (CVI) improves cIoU by \textcolor{teal}{$2.2\%$}, \textcolor{teal}{$1.5\%$}, and \textcolor{teal}{$1.0\%$} under the $\tfrac{1}{30}$, $\tfrac{1}{16}$, and $\tfrac{1}{8}$ splits, respectively. Output Consistency–driven Pseudo-Labeling (OPCL) adds \textcolor{teal}{$3.4\%$}, \textcolor{teal}{$2.1\%$}, and \textcolor{teal}{$1.4\%$}, while Token-Level Feature Consistency Alignment (TFCA) contributes another \textcolor{teal}{$4.0\%$}, \textcolor{teal}{$2.5\%$}, and \textcolor{teal}{$2.0\%$}. Together, these results demonstrate the complementary benefits of CVI, OPCL, and TFCA.

 
\section{Conclusion}

In this work, we introduce CORA (Consistency-guided Object Relational Alignment), a semi-supervised framework for reasoning-based segmentation. CORA addresses key challenges in low-label regimes by combining spatially grounded visual instructions with consistency-aware learning strategies. Specifically, our approach integrates conditional visual prompts to enrich supervision on labeled data, output consistency–driven pseudo-label filtering to improve reliability on unlabeled data, and token-level feature alignment to ensure consistency across labeled and unlabeled images. We demonstrate the effectiveness of CORA on two diverse and challenging domains: a) Cityscapes dataset, covering real-world autonomous driving scenarios, and b) PanNuke dataset, a complex biomedical dataset requiring fine-grained reasoning and localization. In both cases, CORA achieves strong performance under limited supervision, outperforming existing baselines. These results demonstrate the effectiveness of consistency-guided learning for reasoning segmentation in low-label regimes.

\textbf{Acknowledgement.} This work was partially supported by the National Science Foundation (IIS-2123920,  IIS-2212046) and NSF's pilot initiative (NAIRR240338), which provided credits to access GPT-3.5 Turbo.


{
    \small
    \bibliographystyle{ieeenat_fullname}
    \bibliography{main}
}

\end{document}